\pdfoutput=1

\PassOptionsToPackage{table}{xcolor}

\documentclass[11pt]{article}

\usepackage[preprint]{acl}

\usepackage{latexsym}

\usepackage{amssymb} 
\usepackage{pifont} 
\usepackage{rotating} 
\usepackage{array}
\usepackage{booktabs}

\newcommand{\cmark}{\ding{51}}%
\newcommand{\xmark}{\ding{55}}%
\newcommand{\greencheck}{\textcolor{green!70!black}{\cmark}}
\newcommand{\redx}{\textcolor{red!70!black}{\xmark}}
\newcommand{\partialcheck}{\textcolor{orange!70!black}{$\sim$}}

\usepackage[T1]{fontenc}

\usepackage[utf8]{inputenc}

\usepackage{microtype}
\usepackage{makecell} 

\usepackage{inconsolata}

\usepackage{times}
\usepackage{amsmath}
\usepackage{graphicx}
\usepackage{url}
\usepackage{booktabs}
\usepackage{multirow}
\usepackage{listings}

\title{GLiNER2: An Efficient Multi-Task Information Extraction System \\with Schema-Driven Interface}

\author{
  Urchade Zaratiana, 
  Gil Pasternak,
  Oliver Boyd
  \\
  \textbf{George Hurn-Maloney, 
  Ash Lewis}
  \\
  Fastino AI \\
  \texttt{\{uz,gil,o8,g,ash\}@fastino.ai}
}

\begin{document}
\maketitle
\begin{abstract}
Information extraction (IE) is fundamental to numerous NLP applications, yet existing solutions often require specialized models for different tasks or rely on computationally expensive large language models. We present GLiNER2, a unified framework that enhances the original GLiNER architecture to support named entity recognition, text classification, and hierarchical structured data extraction within a single efficient model. Built pretrained transformer encoder architecture, GLiNER2 maintains CPU efficiency and compact size while introducing multi-task composition through an intuitive schema-based interface. Our experiments demonstrate competitive performance across extraction and classification tasks with substantial improvements in deployment accessibility compared to LLM-based alternatives. We release GLiNER2 as an open-source pip-installable library with pre-trained models and documentation at \href{https://github.com/fastino-ai/GLiNER2}{github.com/fastino-ai/GLiNER2}.

\end{abstract}

\section{Introduction}
Information extraction (IE) \citep{okurowski-1993-information,weischedel-etal-1996-progress} represents one of the most fundamental and practically important tasks in natural language processing, involving the identification and extraction of structured information from unstructured text. While large language models \citep{gpt4} have demonstrated remarkable capabilities across various IE tasks \citep{wang-etal-2025-gpt,han2024empirical}, their deployment presents significant practical challenges that limit their accessibility and adoption. Smaller models \citep{touvron2023llama2openfoundation, jiang2023mistral7b, yang2025qwen3technicalreport} (eg. 
\textit{Llama-2-7b}) require GPU acceleration to achieve reasonable inference speeds, making CPU-based deployment prohibitively slow for production use. The availability of GPU resources can be a barrier for organizations and researchers operating under resource constraints.

\begin{table*}[h]
\centering
\small
\begin{tabular}{lcccc}
\toprule
\textbf{Characteristic} & \textbf{GLiNER} & \textbf{GLiNER2} & \textbf{Open LLMs} & \textbf{Closed LLMs} \\
\midrule
\multicolumn{5}{l}{\textit{\textbf{Features}}} \\
Scope & NER only & Various IE \& Classification & General & General \\
Label description & \redx & \greencheck & \greencheck & \greencheck \\
CPU Deployment & \greencheck & \greencheck & \redx & \redx \\
Privacy Preserving & \greencheck & \greencheck & \greencheck & \redx \\
No API Costs & \greencheck & \greencheck & \greencheck & \redx \\
Fine-tuning Support & \greencheck & \greencheck & \greencheck & \partialcheck \\
\midrule
\multicolumn{5}{l}{\textit{\textbf{Technical Specifications}}} \\
Parameters & 195M & 205M & 7B-175B & Unknown \\
Model Architecture & Encoder & Encoder & Decoder & Decoder \\
Context Length & 512 tokens & 2048 tokens & 2K-1M tokens & 8K-10M tokens \\
\midrule
\multicolumn{5}{l}{\textit{\textbf{Usage \& Licensing}}} \\
License Type & Apache 2.0 & Apache 2.0 & Various & Proprietary \\
Commercial Use & \greencheck & \greencheck & \partialcheck & \greencheck \\
\bottomrule
\end{tabular}
\caption{Comprehensive comparison across system categories. \greencheck = full support, \partialcheck = partial/limited support, \redx = no support.}
\label{tab:comprehensive-comparison}
\end{table*}
\paragraph{} Beyond computational requirements, LLMs present additional deployment challenges. Although API costs have decreased significantly in recent years, they can pose serious privacy and security concerns \citep{Shanmugarasa2025SoKTP,wu2025indepthinvestigationdatacollection}, particularly when processing sensitive data containing personally identifiable information (PII), financial records, or proprietary business information. Many organizations in the healthcare, finance, and government sectors require on-premises deployment to maintain data sovereignty and comply with regulations such as GDPR, HIPAA, or industry-specific compliance requirements \citep{lareo2023llm, zhang2024rightforgotteneralarge, cnil2024ai}. Furthermore, the recurring costs associated with API usage may remain prohibitive for researchers, startups, and practitioners in developing countries, creating inequitable access to advanced NLP capabilities. Our goal is to address these issues specifically in the context of information extraction.

\paragraph{} GLiNER \citep{zaratiana-etal-2024-gliner} was proposed to address these fundamental limitations specifically for named entity recognition (NER) tasks. GLiNER \citep{zaratiana-etal-2024-gliner} introduced a paradigm shift by enabling zero-shot NER using a small transformer encoder architecture trained on diverse, LLM-annotated datasets \citep{zhou2024universalner,bogdanov-etal-2024-nuner}. This approach yielded remarkable results, achieving performance that matched or surpassed contemporary LLMs while running efficiently on standard CPU hardware without requiring GPUs. Furthermore, it maintains a parameter count under 500 million, enabling deployment in edge computing scenarios, resource-constrained environments, and privacy-sensitive applications. GLiNER gained particular traction in the PII redaction domain \citep{gretel2024gliner, presidio_gliner_sample}, where its combination of competitive performance, CPU efficiency, and straightforward local deployment made it an ideal solution for handling sensitive data. 

Following GLiNER's release, several specialized adaptations emerged for different information extraction tasks. GLiREL \citep{boylan-etal-2025-glirel} extended the approach to relation extraction, while GLiClass \citep{gliclass2025} adapted it for zero-shot text classification. Domain-specific variants included GLiNER-BioMed \citep{yazdani2025glinerbiomedsuiteefficientmodels} for biomedical entity recognition, OpenBioNER \citep{cocchieri_openbioner_2025} for lightweight biomedical NER through entity type descriptions, and GLiDRE \citep{armingaud2025glidre} for document-level relation extraction in French. However, each adaptation required \textit{separate model development and deployment}, leading to fragmentation as practitioners needed multiple specialized models for comprehensive information extraction pipelines.

\paragraph{} In this work, we introduce GLiNER2, a Python library that transforms the focused capabilities of its predecessors into a \textit{universal information extraction system}. Rather than requiring separate models like GLiNER\footnote{\url{https://github.com/urchade/GLiNER}}, GLiREL\footnote{\url{https://github.com/jackboyla/GLiREL}}, and GLiClass\footnote{\url{https://huggingface.co/knowledgator/GLiClass}}, GLiNER2 unifies entity recognition, structured extraction, and text classification within a single architecture. GLiNER2 maintains the core efficiency, running on CPU, while dramatically expanding functionality beyond simple NER to support: entity recognition with natural language type descriptions and nested/overlapping spans, document-level classification with configurable single or multi-label outputs, and complex extraction schemas that capture hierarchical structures with parent-child relationships and repeated patterns. The library's Python API allows developers to define extraction schemas declaratively, compose multiple tasks in a single inference call, and deploy models with just a few lines of code. By unifying these capabilities, GLiNER2 replaces multiple specialized models with a \textit{single efficient solution}. GLiNER2 is available through pip installation (\texttt{pip install gliner2}) with pre-trained weights hosted on Hugging Face, and it is released under the \textit{Apache 2.0} license.

\begin{table*}[h]
\centering
\small
\begin{tabular}{@{}llcccccc@{}}
\toprule
\textbf{Dataset} & \textbf{Task Type} & \textbf{\# Labels} & 
\makecell{\textbf{GPT-4o} \\ \citet{gpt4} \\ \textbf{>100B}} & 
\makecell{\textbf{GLiClass} \\ \citet{gliclass2025} \\ \textbf{190M}} & 
\makecell{\textbf{DeBERTa-v3} \\ \citet{laurer_building_2023} \\ \textbf{435M}} & 
\makecell{\textbf{GLiNER2} \\ \textit{Our model} \\ \textbf{205M}} \\
\midrule
SNIPS            & Intent     & 7   & 0.97 & 0.80 & 0.77 & 0.83 \\
Banking77        & Intent     & 77  & 0.78 & 0.21 & 0.42 & 0.70 \\
Amazon Intent    & Intent     & 31  & 0.72 & 0.51 & 0.59 & 0.53 \\
\midrule
SST-2            & Sentiment  & 2   & 0.94 & 0.90 & 0.92 & 0.86 \\
IMDB             & Sentiment  & 2   & 0.95 & 0.92 & 0.89 & 0.87 \\
\midrule
AG News          & Topic      & 4   & 0.85 & 0.68 & 0.68 & 0.74 \\
20 Newsgroups    & Topic      & 20  & 0.68 & 0.36 & 0.54 & 0.49 \\
\midrule
\textbf{Average} & —          & —   & \textit{0.84} & 0.63 & 0.69 & \textbf{0.72} \\
\bottomrule
\end{tabular}
\caption{Zero-shot text classification performance across various benchmarks.}
\label{tab:classification_results}
\end{table*}

\section{System design}

Our architecture builds upon the foundational design principles of the original GLiNER \cite{zaratiana-etal-2024-gliner}, which prompts a pretrained transformer encoder \citep{devlin_bert_2019, he2023debertav} with entity types for zero-shot named entity recognition. We extend this prompting approach to handle more complex schemas that encompass multiple information extraction tasks. The core innovation lies in our unified input formulation that enables diverse extraction tasks through carefully designed prompt templates. The general input format follows:
\begin{equation*}
\texttt{[Task Prompt]} \oplus \texttt{[SEP]} \oplus \texttt{[Input Text]}
\end{equation*}

where $\oplus$ denotes concatenation. The \texttt{[Task Prompt]} specifies what to extract (e.g., entity types like "person, location" or class labels like "positive, negative"), \texttt{[SEP]} is a special separator token, and \texttt{[Input Text]} is the text to be analyzed, which is a sequence of text tokens $x_1, x_2, \ldots, x_N$. Complete task prompt formats for each extraction type are detailed in Appendix~\ref{sec:appendix}.

\paragraph{} GLiNER2 comprises the following tasks:
\begin{itemize}
    \item \textbf{Entity Recognition:} We support entity type descriptions alongside labels, allowing richer semantic understanding through natural language definitions.
    
    \item \textbf{Hierarchical Structure Extraction:} We introduce structured schemas that capture parent-child relationships between entities and their attributes, enabling extraction of complex nested information.
    
    \item \textbf{Text Classification:} We add text classification capabilities with support for both single-label and multi-label, along with label description.
    
    \item \textbf{Task Composition:} Most significantly, we enable composition of multiple extraction tasks within a single forward pass, allowing simultaneous entity recognition, text classification, and structured extraction with shared contextual understanding.
\end{itemize}

This unified approach maintains the efficiency advantages of the original GLiNER while dramatically expanding its capabilities to handle diverse information extraction scenarios. Detailed architectural specifications and mathematical formulations are provided in Appendix \ref{sec:appendix}.

\section{Experiments}

\subsection{Training Data}

We trained our model on 254,334 examples, combining \textit{real-world documents} and \textit{synthetic data}, with balanced coverage of entity recognition, hierarchical extraction, and classification tasks. The real-world set consists of 135,698 documents from \textit{news articles}, \textit{Wikipedia}, \textit{legal texts}, \textit{PubMed abstracts}, and \textit{ArXiv papers}, representing a wide range of writing styles and entity types. All documents were automatically annotated with \textit{GPT-4o} using task-specific prompts and validated for quality. To address gaps and improve robustness, we generated 118,636 \textit{synthetic examples} with \textit{GPT-4o} targeting common business and personal use cases, including \textit{email threads}, \textit{text messages}, \textit{professional documents}, \textit{social media posts}, \textit{transactional data}, and \textit{domain-specific texts}; each synthetic example includes complete annotations for all tasks to support effective multi-task learning. Full dataset statistics and distribution are provided in Appendix~\ref{app:training_data}.

\subsection{Results}

We conducted comprehensive zero-shot evaluations on standard benchmarks for both text classification and named entity recognition to assess the effectiveness of our approach. Hierarchical structure extraction was not evaluated due to the absence of established zero-shot benchmarks for this task type, which we plan to address in future work. Details on all baseline models and evaluation protocols are provided in Appendix~\ref{app:exper}.

\paragraph{} We evaluate zero-shot classification on seven public benchmarks covering sentiment (\textit{SST-2} \citep{socher_recursive_2013}, \textit{IMDB} \citep{maas_learning_2011}), intent (\textit{SNIPS} \citep{coucke2018snipsvoiceplatformembedded}, \textit{Banking77}  \citep{casanueva-etal-2020-efficient}, \textit{Amazon-Intent} \citep{fitzgerald2022massive1mexamplemultilingualnatural}), and topic classification (\textit{AG News} \citep{zhang2016characterlevelconvolutionalnetworkstext}, \textit{20 Newsgroups} \citep{Lang1995NewsWeederLT}). \textbf{GLiNER2 achieves the highest average accuracy among open-source baselines}, outperforming GLiClass on five datasets and DeBERTa-v3 on three. It performs particularly well on intent classification, scoring 0.83 on \textit{SNIPS} and 0.70 on \textit{Banking77}, compared to DeBERTa's 0.77 and 0.42, respectively. On \textit{Amazon-Intent} and \textit{20 Newsgroups}, GLiNER2 trails DeBERTa-v3 slightly (by 6 and 5 points), but GLiNER2 is significantly faster as shown in Table~\ref{tab:latency_ordered}. On sentiment benchmarks, GLiNER2 scores 0.86–0.87, close to DeBERTa-v3's 0.89–0.92 and within 10 points of GPT-4o. While GPT-4o leads across all tasks, this superior performance is expected given its substantially larger scale and extensive pretraining on diverse text corpora. Overall, GLiNER2 offers competitive accuracy across a range of classification settings and consistently closes the gap between task-specific baselines and large proprietary models.

\begin{table}[]
\centering
\small
\begin{tabular}{l|ccc}
\toprule
\textbf{Dataset} & \textbf{GPT-4o} & \textbf{GLiNER-M} & \textbf{GLiNER2} \\
\midrule
AI         & 0.547 & 0.518 & 0.526 \\
Literature & 0.561 & 0.597 & 0.564 \\
Music      & 0.736 & 0.694 & 0.632 \\
Politics   & 0.632 & 0.686 & 0.679 \\
Science    & 0.518 & 0.581 & 0.547 \\
\midrule
\textbf{Average} & 0.599 & 0.615 & 0.590 \\
\bottomrule
\end{tabular}
\caption{Zero-shot F1 scores on CrossNER benchmark.}
\label{tab:crossner_gliner2}
\end{table}

\paragraph{} For NER evaluation, we use the \textit{CrossNER} benchmark \citep{liu2020crossnerevaluatingcrossdomainnamed}, which measures zero-shot generalization across five specialized domains: \textit{AI}, \textit{Literature}, \textit{Music}, \textit{Politics}, and \textit{Science}. As shown in Table~\ref{tab:crossner_gliner2}, \textbf{GLiNER2 closely matches GPT-4o in overall F1 score} (0.590 vs.\ 0.599) and achieves higher scores in \textit{AI} (0.526 vs.\ 0.547) and \textit{Literature} (0.564 vs.\ 0.561). While GLiNER2 trails GLiNER-M in categories like \textit{Science} and \textit{Music}, it maintains strong performance in \textit{Politics} (0.679), suggesting robustness across diverse entity types. Considering that GLiNER2 is a general-purpose model supporting multiple tasks, this level of NER performance with only modest drop-offs compared to a dedicated entity recognition system, demonstrates the effectiveness of our unified architecture.

\subsection{Efficiency}

We evaluate GLiNER2's computational efficiency by measuring inference latency on text classification tasks across different numbers of labels. All models are evaluated on \textit{CPU} except GPT-4o, which uses the \textit{OpenAI API}. Table~\ref{tab:latency_ordered} presents latency measurements in milliseconds for varying numbers of classification labels. GLiNER2 demonstrates \textbf{strong computational efficiency}, achieving latency comparable to GLiClass while providing significantly better performance than DeBERTa-based zero-shot classification. The key advantage becomes evident when comparing against DeBERTa, which performs a \textit{separate forward pass for each label}, resulting in \textit{linear scaling} with the number of labels and substantially higher latency (\textbf{6.8× slower} with 20 labels). In contrast, GLiNER2 processes \textit{all labels simultaneously} in a single forward pass, maintaining consistent performance regardless of label count. Both GLiNER2 and GLiClass achieve approximately \textbf{2.6× speedup over GPT-4o} while running on standard CPU hardware, demonstrating the practical advantages of compact, specialized models for production deployment scenarios where \textit{latency and computational resources} are critical considerations.

\begin{table}[ht]
\centering
\small
\begin{tabular}{@{}r|cccc@{}}
\toprule
\textbf{\#Labels} & \textbf{GPT‑4o} & \textbf{DeBERTa} & \textbf{GLiClass} & \textbf{GLiNER2} \\
\midrule
5   & 358   & 1714  & \textbf{137} & \textbf{130} \\
10  & 382   & 3404  & \textbf{131} & \textbf{132} \\
20  & 425   & 6758  & \textbf{140} & \textbf{163} \\
50  & 463   & 16897 & \textbf{190} & \textbf{208} \\
\midrule
\textbf{Speedup} & 1.00× & 0.10× & 2.75× & 2.62× \\
\bottomrule
\end{tabular}
\caption{CPU Latency (ms) comparison for text classification with varying number of labels.}
\label{tab:latency_ordered}
\end{table}

\section{Artifacts}

\subsection{Python Package}
We provide a Python package that makes GLiNER2 accessible through an intuitive API. The \texttt{gliner2} library can be easily installed via pip and provides seamless integration with the Hugging Face ecosystem for model distribution and loading. The model can be loaded using the standard \texttt{.from\_pretrained} method, with weights hosted on Hugging Face Hub for convenient access.

\begin{figure}[h]
    \centering
\includegraphics[width=1\columnwidth]{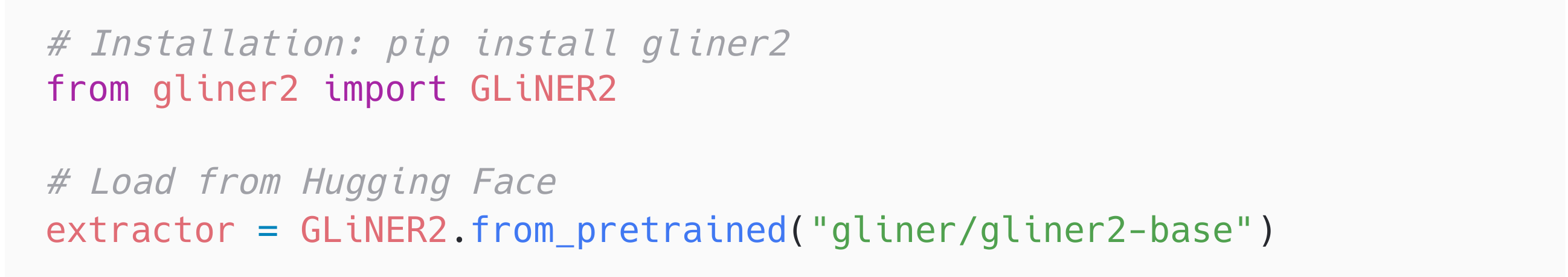}
    \vspace{-1.8em}
    \caption{Model loading.}
        \label{fig:load}
\end{figure}

\begin{figure}[h]
    \centering
\includegraphics[width=1\columnwidth]{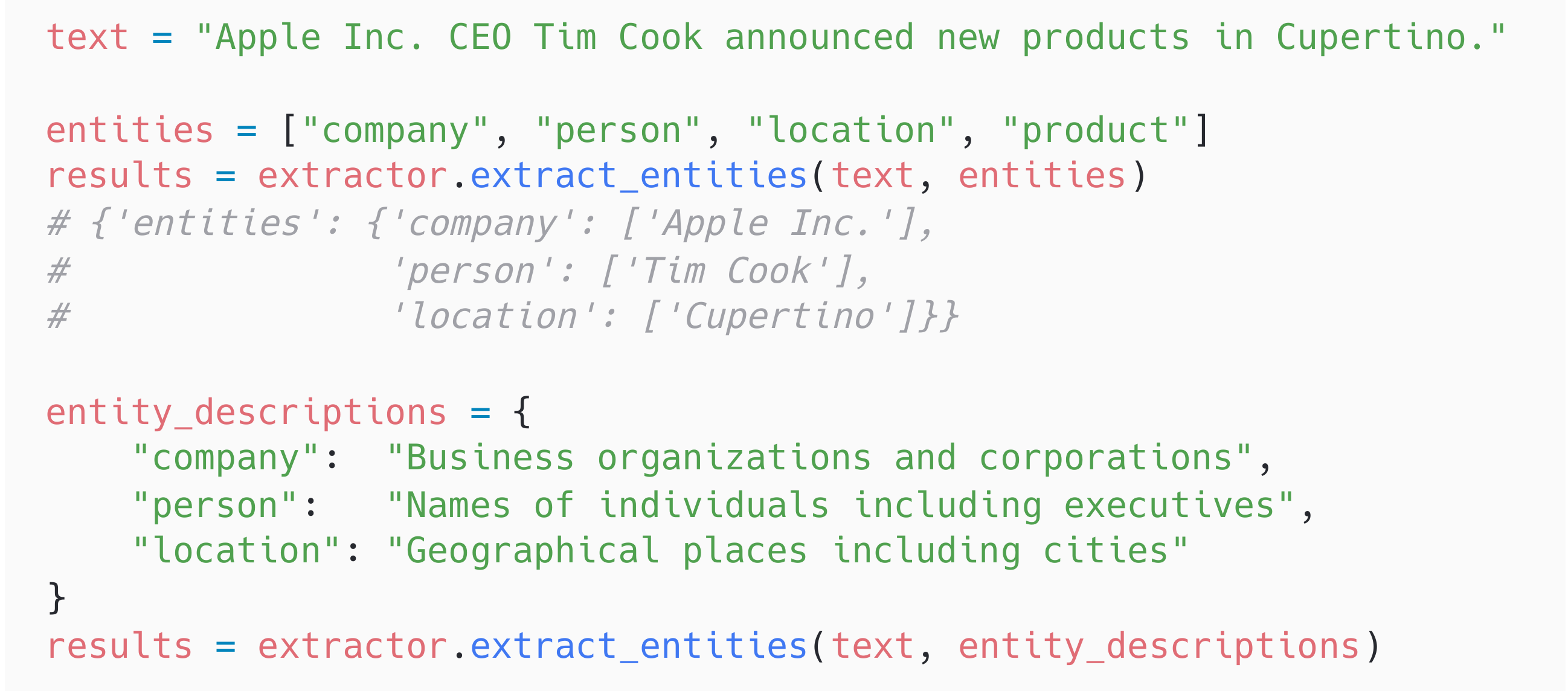}
    \vspace{-1.8em}
    \caption{GLiNER2 for Named Entity Recognition with simple and enhanced approaches}
    \label{fig:ner}
\end{figure}

\paragraph{Named Entity Recognition}
Named entity recognition can be performed through multiple approaches to accommodate different use cases and complexity levels. The simplest method requires only the input text and a list of target entity types. Moreover, users can provide entity types with natural language descriptions using a dictionary format, where keys represent entity types and values contain descriptive text that helps the model better understand the extraction target. The process and various usage patterns are illustrated in Figure \ref{fig:ner}.

\paragraph{Hierarchical Structure Extraction}

\begin{figure}[!h]
\centering
\includegraphics[width=1\columnwidth]{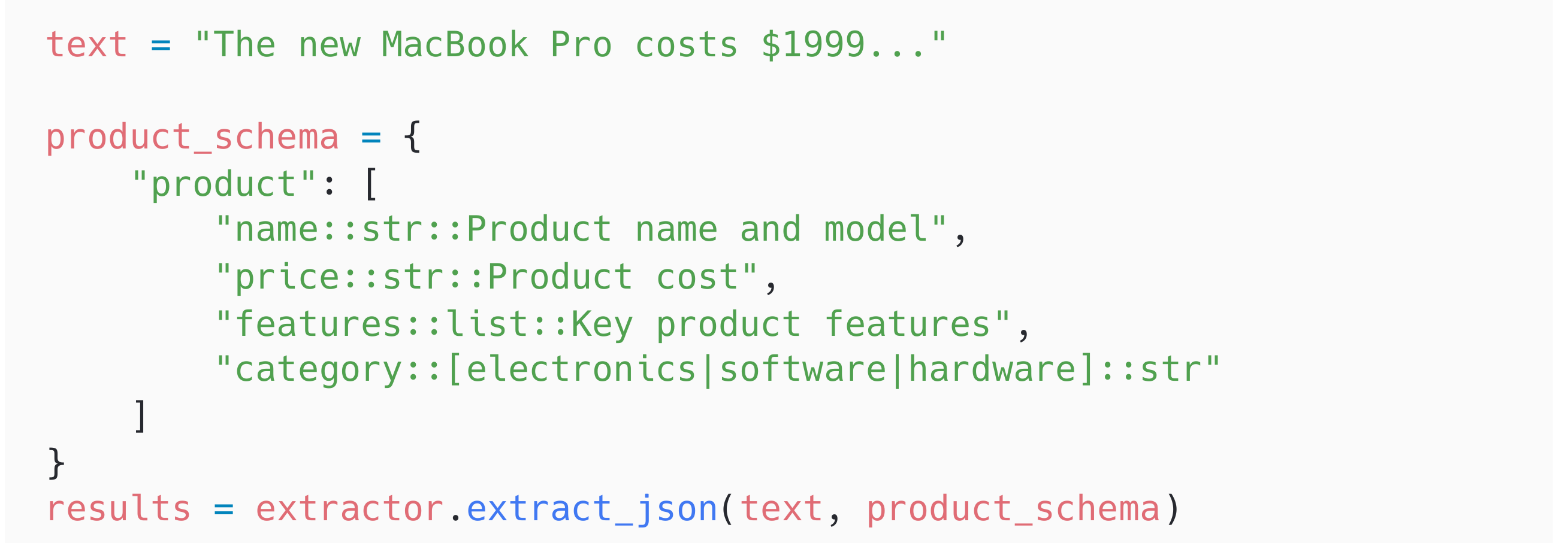}
\vspace{-1.8em}
\caption{Hierarchical structure extraction with field constraints and descriptions}
\label{fig:hier}
\end{figure}

Hierarchical structure extraction is performed by defining a schema as shown in Figure \ref{fig:hier}. The schema defines a parent entity (termed a \textit{structure}) containing multiple child fields using GLiNER2's field specification syntax. Each field follows the pattern \texttt{field\_name::type::description}, where \texttt{type} specifies either \texttt{str} for single values or \texttt{list} for multiple values. Fields may incorporate choice constraints through the format \texttt{field\_name::[option1|option2]::type}, exemplified by the category field restricted to \textit{electronics}, \textit{software}, or \textit{hardware}.

\begin{figure}[!h]
\centering
\includegraphics[width=1\columnwidth]{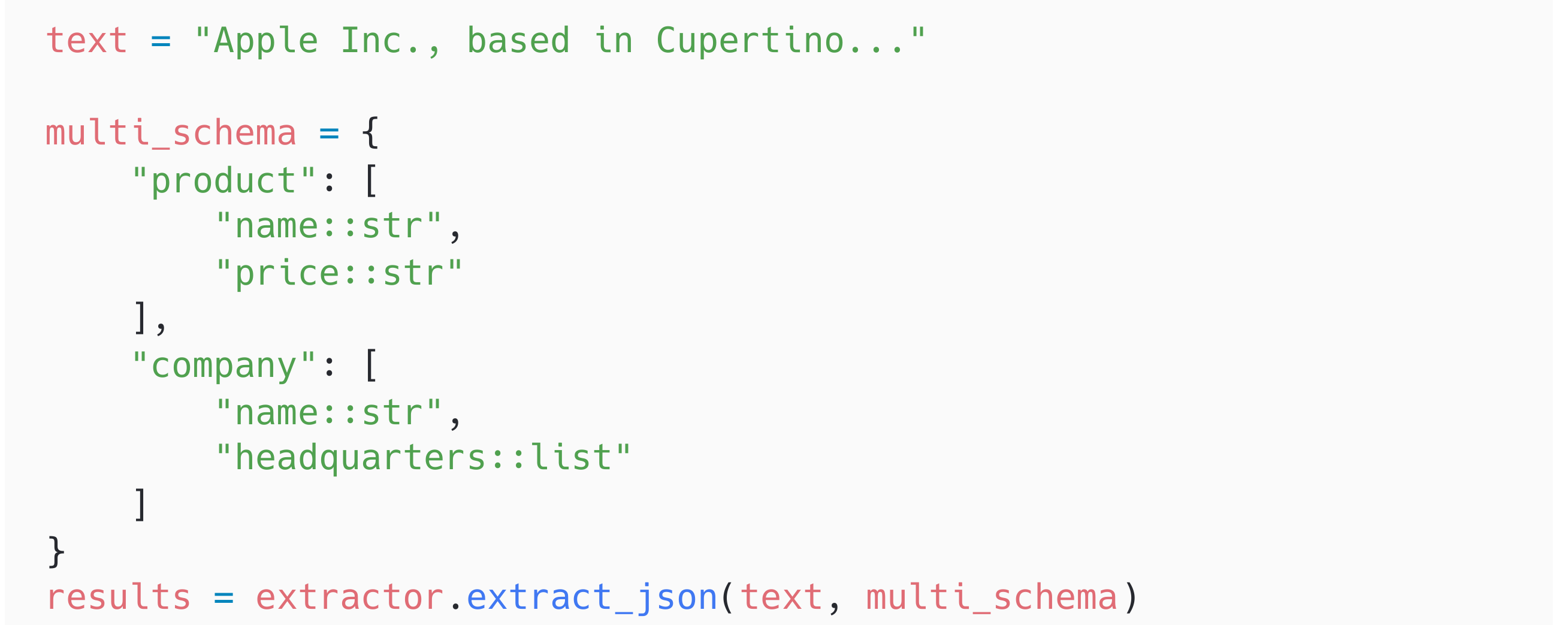}
\vspace{-1.8em}
\caption{Composing multiple hierarchical structures in a single schema}
\label{fig:mulhier}
\end{figure}

The framework supports multiple structures within a single schema for complex extraction scenarios. For instance, Figure~\ref{fig:mulhier} shows how users can define two structures, \textit{product} and \textit{company}, in a single query. This enables simultaneous extraction of product details (\textit{name} and \textit{price}) alongside company information (\textit{name} and \textit{headquarters}), all processed efficiently in a single forward pass.

\paragraph{Text Classification}

\begin{figure}[h]
    \centering
\includegraphics[width=1\columnwidth]{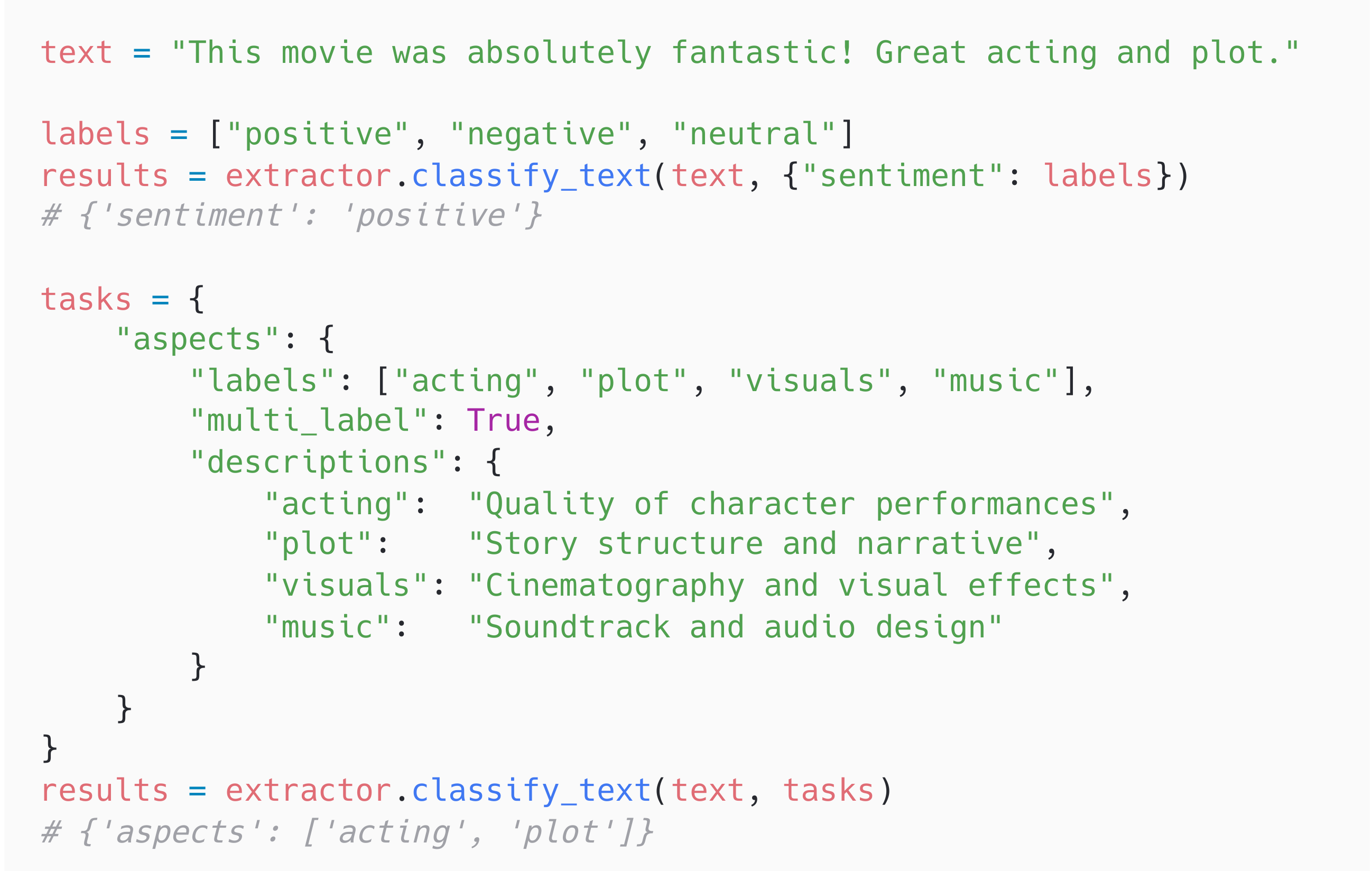}
    \vspace{-1.8em}
    \caption{Text classification with simple and advanced configuration options}
    \label{fig:class}
\end{figure}

Like NER, text classification functionality provides both streamlined and highly customizable interfaces to accommodate various application requirements. For quick deployment, users need only provide the input text and a dictionary mapping task names (e.g., "sentiment") to lists of classification labels, as shown in the first example of Figure \ref{fig:class}. For more sophisticated applications, the library supports extensive customization options including label descriptions and multi-label classification capabilities. When multi-label classification is enabled, the model applies sigmoid activation to allow multiple simultaneous label assignments, while single-label tasks use softmax normalization for mutually exclusive predictions.

\begin{figure}[h]
    \centering
\includegraphics[width=1\columnwidth]{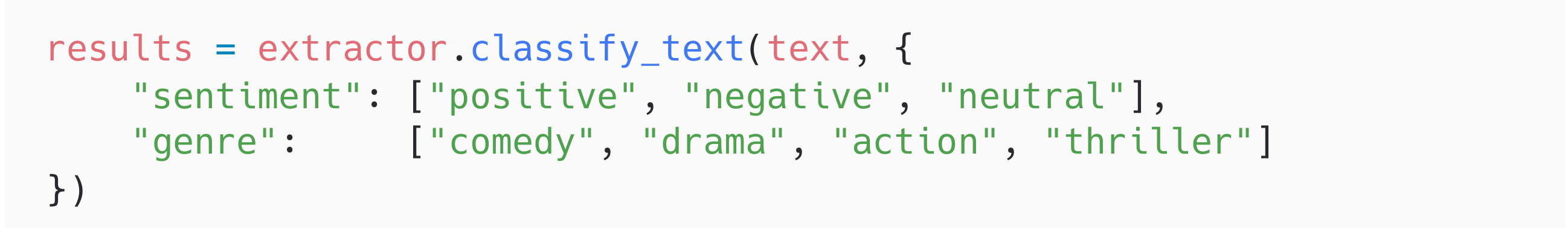}
    \vspace{-1.8em}
    \caption{Simultaneous multi-task classification.}
    \label{fig:mulclass}
\end{figure}

The library supports multiple classification tasks within a single call, as demonstrated in Figure \ref{fig:mulclass}. Each classification task can be independently customized with features such as label descriptions and multi-label settings.

\begin{figure}
    \centering
\includegraphics[width=1\columnwidth]{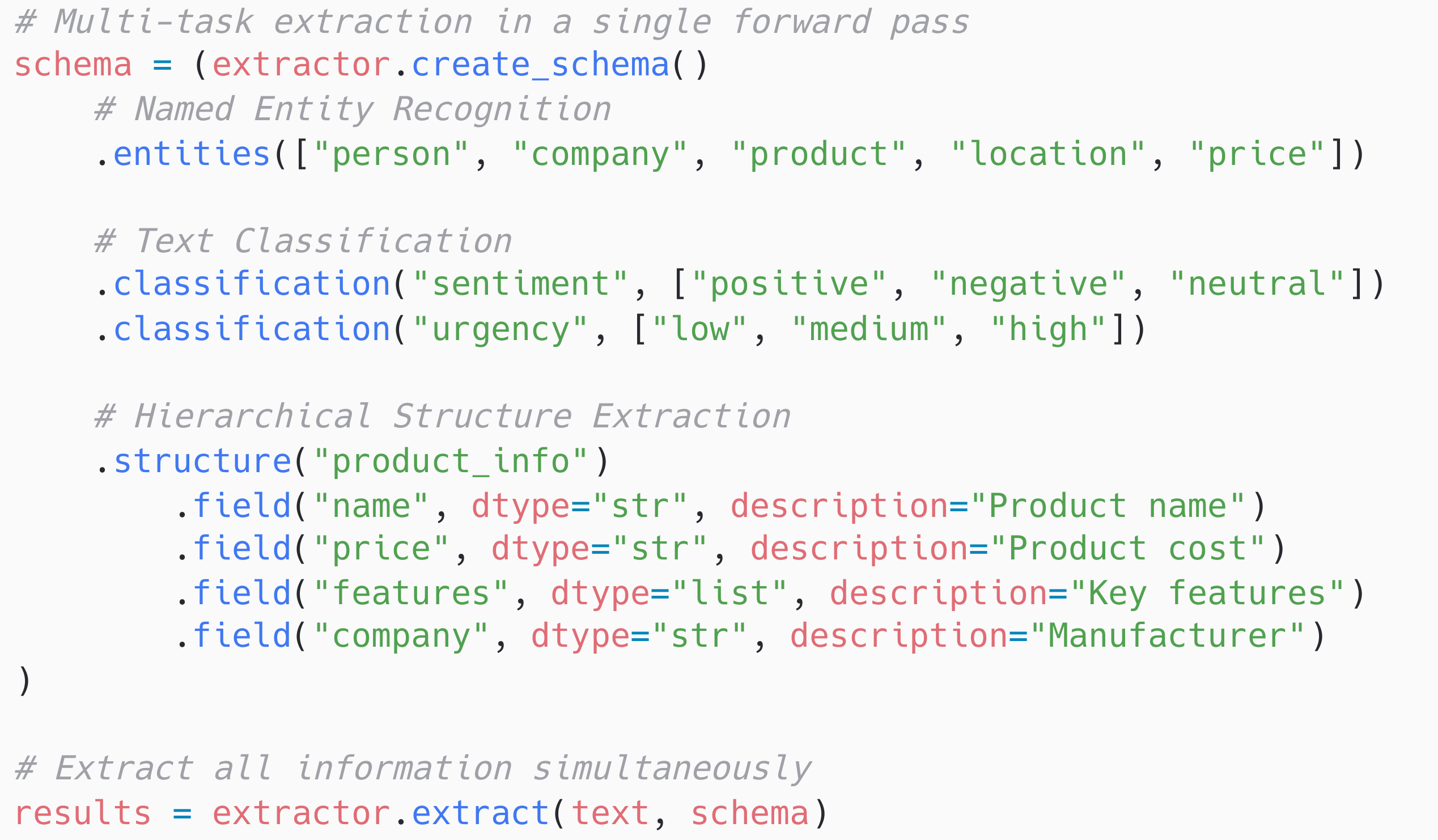}
    \vspace{-1.8em}
    \caption{Comprehensive task composition combining all extraction types}
    \label{fig:taskcom}
\end{figure}

\paragraph{Task Composition}
A key feature of the library is its ability to efficiently compose multiple extraction tasks within a single unified framework. Figure \ref{fig:taskcom} demonstrates how to construct a comprehensive schema that combines entity recognition, text classification, and structured extraction seamlessly in one inference call. 

\begin{figure*}
    \centering
\includegraphics[width=1\textwidth]{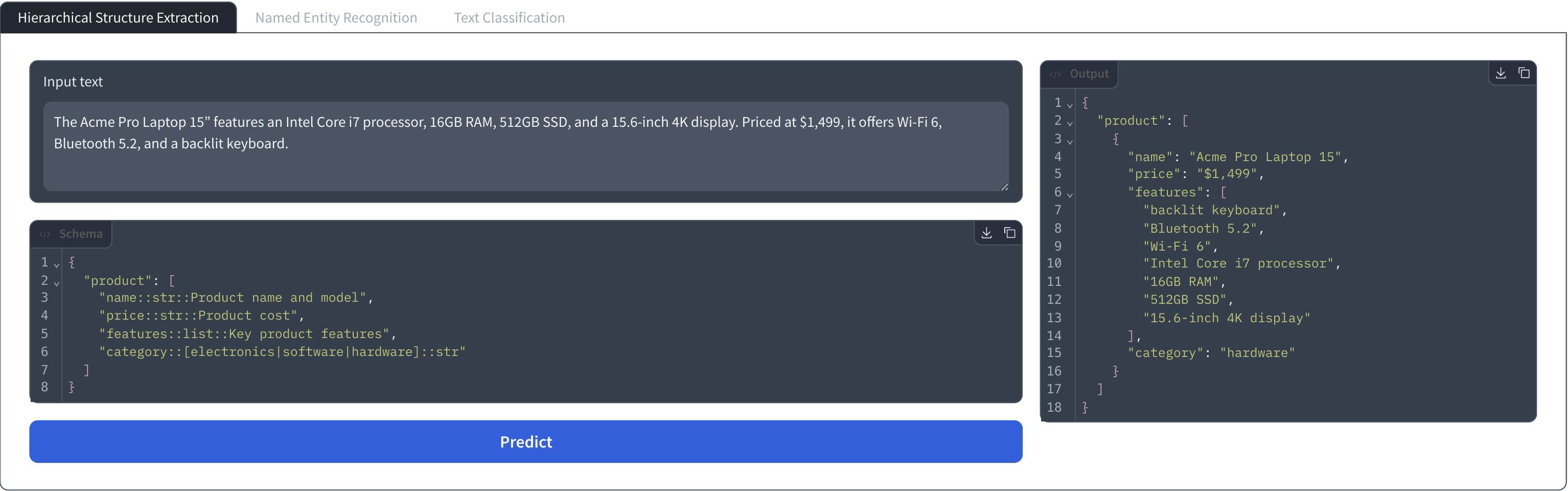}
    \vspace{-1.8em}
\caption{GLiNER2 Gradio demo interface showing hierarchical structure extraction.}    \label{fig:screenshot}
\end{figure*}

\subsection{Interactive Gradio Demo}

We provide a web-based demonstration interface that allows users to interact with GLiNER2 without writing code. The demo enables real-time experimentation with entity types, classification labels, descriptions and other parameters. The interface consists of three tabs corresponding to GLiNER2's core capabilities. Figure \ref{fig:screenshot} shows the hierarchical structure extraction tab, where users can define schemas with multiple fields and data types to extract structured information from text.

\section{Related Work}
Several frameworks have addressed information extraction tasks across different domains and approaches.
\paragraph{Traditional NLP Libraries:} spaCy \citep{honnibal2020spacy}, Stanford CoreNLP \citep{manning_stanford_2014}, Stanza \citep{qi_stanza_2020} provide comprehensive toolkits for named entity recognition, part-of-speech tagging, and dependency parsing. However, these frameworks require separate models for each task and lack unified architectures, and often does not generalize to unseen labels.
\paragraph{LLM-based Extraction:} XNLP \citep{fei_xnlp_2024} demonstrated using large language models for diverse IE tasks through prompting strategies, while NuExtract \citep{numind2024nuextract} focused on fine-tuning for JSON extraction. These approaches achieve strong performance but require significant computational resources and GPU inference.
\paragraph{Encoder-based Approaches:} GLiNER \citep{zaratiana-etal-2024-gliner} introduced an efficient paradigm leveraging pretrained encoders fine-tuned on synthetic data for zero-shot named entity recognition, achieving fast CPU inference with competitive accuracy. This approach inspired subsequent work including GLiClass \citep{gliclass2025} for text classification and GLiREL \citep{boylan-etal-2025-glirel} for zero-shot relation extraction. GLiNER2 extends this line of work by integrating multiple tasks within a single efficient framework, enabling multi-task composition while maintaining the computational advantages of compact encoder-based models.

\section{Conclusion}

We presented GLiNER2, which unifies entity recognition, text classification, and hierarchical extraction in a single CPU-efficient model. Unlike existing approaches requiring separate models per task, GLiNER2 enables multi-task extraction through declarative schemas while maintaining under 500M parameters for practical deployment. We release GLiNER2 as an open-source Python library under Apache 2.0 license, with pre-trained weights on Hugging Face. By combining efficiency with versatility, we hope our library makes advanced information extraction accessible for both research and production use.

\bibliography{custom}

\newpage

\appendix

\section{Architecture Details}
\label{sec:appendix}

\paragraph{Special Token Vocabulary}
Our architecture employs a set of learned special tokens, each serving a specific semantic role:

\begin{itemize}
   \item \textcolor{blue}{\texttt{[P]}} (Prompt): Marks the beginning of task specifications, signaling the model to interpret subsequent tokens as task metadata
   \item \textcolor{orange}{\texttt{[E]}} (Entity): Precedes each entity type in NER tasks to create distinct embeddings for entity categories
   \item \textcolor{green}{\texttt{[C]}} (Child/Component): Indicates attribute fields in hierarchical structures and establishes parent-child relationships
   \item \textcolor{red}{\texttt{[L]}} (Label): Denotes classification options, with each label receiving a unique embedding for scoring
   \item \textcolor{purple}{\texttt{[SEP]}} (Separator): Delimits different input segments to prevent information leakage between task specifications and content.
\end{itemize}

These tokens are randomly initialized and learned during training, allowing the model to develop task-specific representations.

\paragraph{Named Entity Recognition}

NER tasks follow the input format:
\begin{flushleft}
\texttt{\textcolor{blue}{[P]} entities (\textcolor{orange}{[E]} $e_1$ \textcolor{orange}{[E]} $e_2$ ... \textcolor{orange}{[E]} $e_n$) \textcolor{purple}{[SEP]} $x_1, x_2, \ldots, x_N$}
\end{flushleft}

During extraction, each \textcolor{orange}{\texttt{[E]}} token generates an embedding representing its entity type. The model creates representations for all possible text spans up to a maximum width, then computes matching scores between span-entity pairs using:
\begin{equation}
\text{score}(s_i, e_j) = \text{sim}(\mathbf{h}_{s_i}, \mathbf{h}_{e_j})
\end{equation}
where $\mathbf{h}_{s_i}$ is the span representation, $\mathbf{h}_{e_j}$ is the entity type embedding, and $\text{sim}(\cdot, \cdot)$ is the dot product with sigmoid activation.

For example, given \texttt{\textcolor{blue}{[P]} entities (\textcolor{orange}{[E]} person \textcolor{orange}{[E]} location)} and text \textit{"John works in Paris"}, all span candidates (e.g., \textit{"John"}, \textit{"works"}, \textit{"Paris"}, \textit{"works in"}) are scored against the entity type embeddings (i.e., representations of each \textcolor{orange}{[E]} token). Spans with a predicted probability above 0.5 for any entity type are selected as entities.

\paragraph{Hierarchical Structure Extraction}

Hierarchical extraction uses the format:
\begin{flushleft}
\texttt{\textcolor{blue}{[P]} \textit{parent} (\textcolor{green}{[C]} $a_1$ \textcolor{green}{[C]} $a_2$ ... \textcolor{green}{[C]} $a_m$) \textcolor{purple}{[SEP]} $x_1, x_2, \ldots, x_N$}
\end{flushleft}

The process operates in two stages. First, an MLP processes the \textcolor{blue}{\texttt{[P]}} token embedding to predict the number $K$ of parent entity instances in the text. This MLP performs 20-class classification (for counts 0-19), trained using the ground-truth instance counts during training. Then, the model generates $K$ distinct representations for each attribute by conditioning the \textcolor{green}{\texttt{[C]}} token embeddings on learned occurrence ID embeddings. Specifically, for each instance $k \in \{1, ..., K\}$, the model combines the base \textcolor{green}{\texttt{[C]}} embeddings with occurrence-specific embeddings learned during training, producing unique representations for each instance-attribute pair. These $K \times m$ representations are matched against text spans using the same scoring mechanism as NER, ensuring each instance maintains separate attribute values. Consider the structured extraction task:
\begin{center}
\texttt{\textcolor{blue}{[P]} product (\textcolor{green}{[C]} name \textcolor{green}{[C]} price)}
\end{center}

Given input text: \textit{"iPhone costs \$999. Galaxy is \$899."} the model processes this in three steps:

\begin{enumerate}
    \item \textbf{Count Prediction}: The MLP count predictor processes the \textcolor{blue}{\texttt{[P]}} token embedding and outputs $K=2$, indicating two product instances are present in the text.
    
    \item \textbf{Representation Generation}: The count embedding layer generates $K$ sets of conditioned representations for each attribute field. This produces two distinct embeddings for \textcolor{green}{\texttt{[C]} name} and two for \textcolor{green}{\texttt{[C]} price}, with each pair corresponding to one product instance.
    
    \item \textbf{Span Extraction}: Each conditioned representation computes similarity scores with all possible text spans as for NER. The model selects the highest-scoring spans for each field while maintaining instance coherence:
    \begin{itemize}
        \item Instance 1: \{name: "iPhone", price: "\$999"\}
        \item Instance 2: \{name: "Galaxy", price: "\$899"\}
    \end{itemize}
\end{enumerate}

This parallel processing enables efficient extraction of multiple structured entities while preserving the semantic relationships between fields within each instance.

\paragraph{Text Classification}

Classification tasks use the format:
\begin{flushleft}
\texttt{\textcolor{blue}{[P]} \textit{task} (\textcolor{red}{[L]} $\ell_1$ \textcolor{red}{[L]} $\ell_2$ ... \textcolor{red}{[L]} $\ell_k$) \textcolor{purple}{[SEP]} $x_1, x_2, \ldots, x_N$}
\end{flushleft}

Each \textcolor{red}{\texttt{[L]}} token produces a label-specific embedding that is refined through a classification head. Specifically, for each label $\ell_i$, the model computes:
\begin{equation}
\text{logit}_i = \text{MLP}(\mathbf{h}_{\ell_i})
\end{equation}
where $\mathbf{h}_{\ell_i}$ is the contextualized embedding from the \textcolor{red}{\texttt{[L]}} token for label $\ell_i$, and MLP is a multi-layer perceptron that projects these embeddings to scalar logits representing label-text compatibility. Single-label tasks apply softmax over all logits to select the highest-probability label, while multi-label scenarios use sigmoid activation on each logit independently. Consider the text classification task:
\begin{flushleft}
\texttt{\textcolor{blue}{[P]} sentiment (\textcolor{red}{[L]} positive \textcolor{red}{[L]} negative \textcolor{red}{[L]} neutral)}
\end{flushleft}

Given input text: \textit{"This movie is amazing!"}. The model processes this in three steps:

\begin{enumerate}
    \item \textbf{Label Embedding Generation}: Each \textcolor{red}{\texttt{[L]}} token creates a distinct embedding for its corresponding label (\textit{positive}, \textit{negative}, \textit{neutral}).
    
    \item \textbf{Classification Head}: The label embeddings are projected through an MLP to produce classification logits, which are then normalized using softmax activation for single-label prediction.
    
    \item \textbf{Label Selection}: The model selects the highest-scoring label, predicting \textit{"positive"} for the given input text.
\end{enumerate}

For multi-label scenarios, sigmoid activation replaces softmax, allowing multiple labels to be selected simultaneously.

\paragraph{Task Composition}

Multiple tasks can be composed for efficient multi-task inference using:
\begin{flushleft}
\texttt{[Task$_1$] $\oplus$ \textcolor{purple}{[SEP]} $\oplus$ [Task$_2$] $\oplus$ ... $\oplus$ \textcolor{purple}{[SEP]} $\oplus$ [$x_1, x_2, \ldots, x_N$]}
\end{flushleft}

This enables simultaneous execution of multiple extraction tasks in a single forward pass. For instance, combining NER and sentiment classification on "Steve Jobs loved the iPhone" extracts entities \{person: ["Steve Jobs"], product: ["iPhone"]\} and sentiment "positive" in one computation, improving efficiency over separate model runs.

\section{Experimental setup}
\label{app:exper}
\paragraph{Baselines} We evaluate our approach against several strong baselines. As an upper bound, we use GPT-4o across all tasks. For classification tasks, we compare against two state-of-the-art open-source models with comparable parameter counts: (1) GLiClass (\textit{knowledgator/gliclass-base-v1.0}) \citep{gliclass2025}, a classification-specific adaptation of GLiNER, and (2) DeBERTa-v3-base-zeroshot (\textit{MoritzLaurer/deberta-v3-large-zeroshot-v2.0}) \citep{laurer_building_2023}, the de facto standard for zero-shot classification on Hugging Face. For NER tasks, we use GLiNER-M performance as reported in \citet{zaratiana-etal-2024-gliner}, which represents the current state-of-the-art for generalist entity recognition.

\paragraph{Hyperparameters} We train our model for 5 epochs using the AdamW optimizer with differential learning rates: $2 \times 10^{-5}$ for task-specific layers and $1 \times 10^{-5}$ for the encoder backbone. This differential approach allows fine-tuning of pretrained representations while enabling faster adaptation of task-specific components. We apply weight decay of 0.01 for regularization and gradient clipping at 1.0 to ensure training stability. The learning rate schedule includes 1,000 warmup steps with linear scaling. Table~\ref{tab:hyperparameters} summarizes the training configuration.

\begin{table}[h]
\centering
\begin{tabular}{ll}
\toprule
\textbf{Hyperparameter} & \textbf{Value} \\
\midrule
Epochs & 5 \\
Optimizer & AdamW \\
Learning rate (backbone) & $1 \times 10^{-5}$ \\
Learning rate (task layers) & $2 \times 10^{-5}$ \\
Weight decay & 0.01 \\
Warmup steps & 1,000 \\
Gradient clipping & 1.0 \\
\bottomrule
\end{tabular}
\caption{Training hyperparameters used across all experiments.}
\label{tab:hyperparameters}
\end{table}

\subsection{Training Data}
\label{app:training_data}

Our training dataset comprises 254,334 examples combining real-world texts and synthetic data. Table~\ref{tab:data_stats} shows the distribution across domains.

\begin{table}[h]
\centering
\begin{tabular}{lr}
\toprule
\textbf{Domain} & \textbf{Count} \\
\midrule
\multicolumn{2}{l}{\textit{Real-world Data}} \\
\quad Law       & 19,798 \\
\quad PubMed    & 16,400 \\
\quad Wikipedia & 17,909 \\
\quad ArXiv     & 7,135 \\
\quad News      & 74,456 \\
\multicolumn{2}{l}{\textit{Synthetic Data}} \\
\quad Mixed Domains & 118,636 \\
\midrule
Total & 254,334 \\
\bottomrule
\end{tabular}
\caption{Training data distribution across domains.}
\label{tab:data_stats}
\end{table}

Real-world data (135,698 examples) was collected from news articles, Wikipedia, legal documents, ArXiv papers, and PubMed abstracts. All texts were annotated using GPT-4o for entity recognition, hierarchical extraction, and classification tasks. Synthetic data (118,636 examples) was generated using GPT-4o to cover practical use cases including emails, text messages, resumes, social media posts, e-commerce orders, banking records, and sports commentary. Each example includes annotations for all applicable task types.

\end{document}